\documentclass[sn-mathphys]{sn-jnl}

\jyear{2021}%

\theoremstyle{thmstyleone}%
\usepackage{float}
\usepackage{multirow}
\usepackage{booktabs}
\usepackage{graphicx}
\usepackage{array}
\usepackage{adjustbox}
\usepackage{tabularx}
\usepackage{amsmath}
\theoremstyle{thmstyletwo}%

\theoremstyle{thmstylethree}%

\begin{document}

\title[Domain Adaptation of Transformer-Based Models]{Domain Adaptation of Transformer-Based Models Using Unlabeled Data for Relevance and Polarity Classification of German Customer Feedback}


\author[1]{\fnm{Ahmad} \sur{Idrissi-Yaghir}}\email{ahmad.idrissi-yaghir@fh-dortmund.de}

\author[1,2]{\fnm{Henning} \sur{Schäfer}}\email{henning.schaefer@uk-essen.de}

\author[1]{\fnm{Nadja} \sur{Bauer}}\email{nadja.bauer@fh-dortmund.de}

\author*[1,3]{\fnm{Christoph M.} \sur{Friedrich}}\email{christoph.friedrich@fh-dortmund.de}

\affil*[1]{\orgdiv{Department of Computer Science}, \orgname{University of Applied Sciences and Arts Dortmund (FHDO)}, \orgaddress{\street{Emil-Figge Str. 42}, \city{Dortmund}, \postcode{44227}, \country{Germany}}}

\affil[2]{\orgdiv{Institute for Transfusion Medicine}, \orgname{University Hospital Essen}, \orgaddress{\street{Hufelandstraße 55}, \city{Essen}, \postcode{45147}, \country{Germany}}}

\affil[3]{\orgdiv{Institute for Medical Informatics}, \orgname{Biometry and Epidemiology (IMIBE), University Hospital Essen}, \orgaddress{\street{Hufelandstraße 55}, \city{Essen}, \postcode{45147}, \country{Germany}}}

\abstract{Understanding customer feedback is becoming a necessity for companies to identify problems and improve their products and services. Text classification and sentiment analysis can play a major role in analyzing this data by using a variety of machine and deep learning approaches.
In this work, different transformer-based models are utilized to explore how efficient these models are when working with a German customer feedback dataset. In addition, these pre-trained models are further analyzed to determine if adapting them to a specific domain using unlabeled data can yield better results than off-the-shelf pre-trained models. To evaluate the models, two downstream tasks from the GermEval 2017 are considered. The experimental results show that transformer-based models can reach significant improvements compared to a fastText baseline and outperform the published scores and previous models. For the subtask \textit{Relevance Classification}, the best models achieve a micro-averaged $F1$-Score of 96.1~\% on the first test set and 95.9~\% on the second one, and a score of 85.1~\% and 85.3~\% for the subtask \textit{Polarity Classification}.}




\maketitle

\section{Introduction}
\label{sec1}
For many companies, customer feedback is a valuable source of information to determine customer satisfaction and opinions about products and services. This feedback enables companies to respond to complaints or requests, which is essential for their success and its continuity. With a growing customer base, the amount of generated feedback becomes unmanageable and challenging, and a manual analysis of the data becomes very time-consuming and in some cases almost impossible. Additionally, feedback is often not directly addressed to the company, but is found on various online platforms and social media, where the relevancy to the company can only be resolved with great effort. These platforms are becoming increasingly important as communication channels between customers and companies. Usually, these platforms contain countless feedback in form of unstructured text data such as posts and tweets. Therefore, automated and intelligent data processing and analysis have become necessary to quickly and efficiently analyze large amounts of customer feedback and extract valuable insights that can be used to improve products and services.

In order to stay up-to-date with customer feedback, companies need to identify relevant feedback and then analyze it.  One approach that has been gaining popularity in recent years is sentiment analysis. This method uses natural language processing (NLP) to automatically analyze text and determine whether the text carries positive, neutral, or negative opinions \citep{2020sent}.

The aim of this paper is to develop models that can automatically classify customer feedback data according to their relevance and polarity. To achieve this objective, different machine learning and deep learning models were employed and compared in terms of their performance.

For this work, the \textit{GermEval} 2017 \citep{germevaltask2017} dataset was chosen to conduct different experiments. The \textit{GermEval 2017 Shared Task on Aspect-based Sentiment in Social Media Customer Feedback} workshop was held to focus on the automatic processing of German language customer feedback, e.g., tweets about \textit{Deutsche Bahn}, a German railroad company. The shared task was divided into four subtasks.

Subtask A: Relevance Classification - This subtask is a binary classification problem and focuses on determining whether a feedback concerns Deutsche Bahn or not.

Subtask B: Document-Level Polarity - This subtask is a multi-class classification problem. According to their polarity (sentiment), the documents should be classified into three categories (positive, negative, or neutral).

Subtask C: Aspect-Level Polarity - For this subtask, all aspects contained in feedback must be identified. Each aspect should then be classified as positive or negative. 

Subtask D: Opinion Target Extraction - The goal of the final subtask is to identify and extract all opinions in a document.

Only the first subtasks (A and B) are taken into consideration in this work. The majority of the used systems utilize transformer-based \citep{Vaswani2017} models, which are fairly new and have shown remarkable results in many tasks across multiple languages. The goal is to investigate whether the subtasks can benefit from these models and improve the micro-averaged $F1$-Score compared to the published scores. In addition, some of the language models are further pre-trained in a second phase using unlabeled domain-specific data with the aim of achieving a domain adaptation. The obtained models are then used to examine their performance on the subtasks. The main contributions of this paper can be summarized as follows:
\begin{itemize}
    \item A comparative study of different approaches and models for the classification of customer feedback data.
    \item An analysis of the effectiveness of domain adaptation and the performance of models after pre-training on domain-specific data.
    \item A discussion on the challenges and limitations of applying deep learning approaches to such tasks.
\end{itemize}

The remainder of this paper is structured as follows. Section ``Related Work'' describes related work, in particular, the used datasets and the results of the GermEval 2017 shared task. Section ``Proposed Method'' details the proposed approach and describes the conducted experiments. Section ''Results and Discussion'' discusses the obtained results and the limitations of sentiment analysis. Finally, Sect. Conclusion presents the conclusion derived from this work. 

\section{Related Work}
\label{sec2}

Text classification is a common problem in NLP and arises in a wide range of applications, such as sentiment analysis, question answering, and topic labeling \citep{Liu2015Sentiment-Analy}. 

Over the years, numerous machine learning approaches have been proposed and applied to tackle the task of sentiment analysis. Earlier, non-deep learning methods, such as support vector machines (SVM) \citep{Cortes_1995}, naïve Bayes (NB) \citep{rish2001empirical} and maximum entropy, were widely used and were considered the state-of-the-art at the time \citep{pang-etal-2002-thumbs}. These are usually paired with text representation approaches, such as bag-of-words (BOW) or term frequency-inverse document frequency (TF-IDF) \citep{tfidf}. With the rise of deep neural networks, new approaches have been developed, starting with the introduction of word embedding models using approaches such as Word2Vec \citep{mikolov2013efficient}, global vectors (GloVe) \citep{Pennington_2014} and fastText \citep{bojanowski-etal-2017-enriching}, which create word vectors with the goal of neighboring similar words in a vector space. Later, new approaches were developed such as embeddings from language models (ELMo) \citep{peters-etal-2018-deep}, which is a deep contextualized word representation model that outperforms Word2Vec. Afterwards, Google introduced the first transformer-based model BERT (bidirectional encoder representations from transformers) \citep{bert_devlin}, which achieved remarkable results in many tasks and started the trend of large transformer-based models. These models are usually pre-trained on large-scale unlabeled task-independent corpora to learn universal language representations. After BERT, models such as RoBERTa \citep{liu2020roberta}, ELECTRA \citep{clark2020electra} were introduced as an improvement over BERT by incorporating new pre-training methods (see Sect. \ref{subsec2}).

Although such transformer-based models outperform previous approaches in various NLP tasks, they might struggle when a task corpus is overly focused on a specific domain \citep{Whang2020}. In this context, Gururangan et al. \citep{gururangan-etal-2020-dont} investigated whether additional pre-training on domain-specific data can be beneficial. The authors propose a domain-adaptive pre-training (DAPT) and a task-adaptive pre-training (TAPT) and conduct experiments on eight classification tasks of four domains to verify the effectiveness of these approaches. For DAPT, unlabeled domain-specific corpora are used, whereas TAPT utilizes the unlabeled task data. The achieved results show that these approaches can improve pre-trained language models, and that the best performance can be reached when these two approaches are used in combination. The approach of adapting transformer-based models to a specific domain already showed success with models such as BioBERT \citep{DBLP:journals/bioinformatics/LeeYKKKSK20}, which was initialized from general BERT and further pre-trained using biomedical text such as PubMed data and was able to improve the results on biomedical tasks. One drawback with these approaches is that even though the model is adapted to the domain, the vocabulary does not contain domain-specific words. This leads to these words being split into multiple sub-words, which can hinder the model learning and degrades its performance. As a possible solution, Souza et al. \citep{10.1007/978-3-030-61377-8_28} applied a language-specific adaption by using a language-specific vocabulary, which was generated over the target language text to train a Brazilian Portuguese model. The model was initialized with multilingual BERT (mBERT) and further pre-trained on Brazilian Portuguese text and was able to improve the performance on a variant of tasks.

\subsection{GermEval 2017 Data}
The data were collected from different internet sources including social media, microblogs, news, and Q\&A sites in the span of one year (May 2015 - June 2016) and were annotated afterward \citep{germevaltask2017}. The obtained dataset\footnote{\url{https://sites.google.com/view/GermEval2017-absa/data?authuser=0} (last access: 19-04-2022)} consists of around 26,000 annotated documents, which were randomly split into 80~\% training, 10~\% development and 10~\% test data. More data was collected from November 2016 to January 2017 to create a further test set. The first test set was called \textbf{synchronic} because it originated from the first data collection, whereas the second was created later on and was therefore named \textbf{diachronic}. The number of documents in each split is shown in Table \ref{documents}. 

\begin{table}[!ht]
\caption{Number of documents in each split}
\centering
\begin{tabular}{cccc}
\toprule
 \textbf{Training} &   \textbf{Development} &  \textbf{Test\_syn} &  \textbf{Test\_dia} \\
\midrule
 20,941 &  2584 &      2566 &      1842 \\
\bottomrule
\end{tabular}
\label{documents}
\end{table}

For the subtasks, data are available in two file formats: tab-separated values (TSV) and extensible markup language (XML). For this work, only the TSV format is used, which contains the following tab-separated fields: document ID (URL), document text, relevance (true or false), document-level polarity (neutral, positive or negative).

The Tables \ref{taskA} and \ref{taskB} show the distribution of each class in the different data splits for the two subtasks.

\begin{table}[!ht]
\caption{Relevance distribution in subtask A data}
\centering
\begin{tabular}{lcc}
\toprule
 \textbf{Dataset} &   \textbf{True} &  \textbf{False} \\
\midrule
     Training &  17,043 ($\approx$ 81.4~\%) &   3898 ($\approx$ 18.6~\%) \\
       Development &   2049 ($\approx$ 79.3~\%) &    535 ($\approx$ 20.7~\%)  \\
 Test\_sync &   2095 ($\approx$ 81.6~\%) &    471 ($\approx$ 18.4~\%)  \\
  Test\_dia &   1547 ($\approx$ 84~\%) &    295 ($\approx$ 16~\%)\\
\bottomrule
\end{tabular}
\label{taskA}
\end{table}

\begin{table}[!ht]
\caption{Sentiment distribution in subtask B data}
\centering
\begin{tabular}{lccc}
\toprule
 \textbf{Dataset} &  \textbf{Neutral} &  \textbf{Negative} &  \textbf{Positive} \\
\midrule
     Training &    14,497 ($\approx$ 69.2~\%) &      5228 ($\approx$ 25~\%) &      1216 ($\approx$ 5.8~\%)\\
       Development &     1812 ($\approx$ 70.1~\%) &       617 ($\approx$ 23.9~\%)  &       155 ($\approx$ 6~\%)  \\
 Test\_sync &     1681 ($\approx$ 65.5~\%) &       780 ($\approx$ 30.4~\%) &       105 ($\approx$ 4.1~\%)  \\
  Test\_dia &     1237 ($\approx$ 67.1~\%)  &       497 ($\approx$ 27~\%) &       108 ($\approx$ 5.9~\%) \\
\bottomrule
\end{tabular}
\label{taskB}
\end{table}

Table \ref{corpus} describes different corpus statistics of the dataset: the count of unique unigrams, bigrams, and trigrams as well as the mean length of the text documents calculated on preprocessed and lowercased data. The applied preprocessing techniques are discussed in Sect. \ref{subsec1}.

\begin{table}[!ht]
\caption{GermEval 2017 corpus statistics}
\centering
\begin{adjustbox}{width=1\textwidth}
\begin{tabular}{ccccc}
\hline
\textbf{Subset} & \textbf{Unigrams} & \textbf{Bigrams} & \textbf{Trigrams} & \textbf{Mean length of documents} \\ \hline
Training & 97,105 & 621,031 & 1,071,487 & 76 \\
Development & 28,038 & 121,234 & 169,832 & 79 \\
Test\_sync & 26,096 & 115,485 & 164,073 & 75 \\
Test\_dia & 18,555 & 78,229 & 108,802 & 68 \\ \hline
\end{tabular}
\end{adjustbox}
\label{corpus}
\end{table}

Examples from the training dataset for subtask A and subtask B are shown below in Table \ref{exampleSent}.

\begin{table}[!ht]
\caption{Examples for document relevance and sentiment}
\centering
\begin{tabularx}{\textwidth}{@{}Xll@{}}
\toprule
 \textbf{Sentence} &  \textbf{Relevance} &  \textbf{Sentiment} \\
\midrule
     Bahn Nr. 1 ist ausgefallen. Nr. 2 kurz vorm Bahnhof stehen geblieben. Bus fast verpasst. Sprint hingelegt. Bus in letzter Sekunde bekommen. (\textbf{Engl.:} Train no. 1 was cancelled. No. 2 stopped shortly before the train station. Almost missed the bus. Sprinted down. Got the bus at the last second.) & True  &      Negative    \\
     \hline
    Ein Land, in dem schnelles kostenloses \#wifi in der Bahn selbstverständlich ist $@$holland\_de \#lekkerradeln (\textbf{Engl.:} A country where fast free \#wifi on the train is a matter of course $@$holland\_de \#lekkerradeln) &     False &       Neutral \\
\bottomrule
\end{tabularx}
\label{exampleSent}
\end{table}

For this work, additional unlabeled German tweets were collected from Twitter with the goal to continue the pre-training of the language models using masked language modeling of one of the described models in Sect. \ref{subsec2}. Similar to the original data, all collected tweets contain the term ``\textit{bahn}'' and originate from the period between January 2017 and October 2021. In German, the used search term can also refer to other words that are not associated with trains or railroads, implying that a portion of the collected data is noisy and does not necessarily belong to the domain of the task. This data were crawled using snscrape\footnote{\url{https://github.com/JustAnotherArchivist/snscrape} (last access: 20-04-2022)} and consists of 1,199,280 tweets. Table \ref{corpus-tweets} lists the statistics of the dataset, which are obtained similarly to Table \ref{corpus}.

\begin{table}[!ht]
\caption{Corpus statistics of the unlabeled tweets collected using the search term ``bahn''}
\centering
\begin{tabular}{cccc}
\hline
\textbf{Unigrams} & \textbf{Bigrams} & \textbf{Trigrams} & \textbf{Mean length of documents} \\ \hline
310,484 & 4,113,318 & 12,313,426 & 21 \\ \hline
\end{tabular}
\label{corpus-tweets}
\end{table}

\subsection{GermEval 2017 Results}

The GermEval 2017 \citep{germevaltask2017} shared task witnessed the participation of 8 teams that used a variety of approaches. All the teams participated in subtask B and 5 of them in subtask A.

 Before utilizing these approaches, the majority of the teams applied a thorough data preprocessing, where they either removed or replaced e.g. URLs, hashtags, handles, and emojis with special tokens. Additionally, some teams used lemmatizers, part-of-speech taggers, and spell checkers. Furthermore, punctuation characters were removed by some teams and kept by others, which was also the case for capitalization.

To evaluate how well these systems perform on the independent test sets, a micro-average $F1$-Score was used. The $F1$-Score, which is short for $F_{\beta=1}$, is generally defined as follows:
\begin{equation}
F_{\beta} = \frac{\left(\beta^{2}+1\right) \times \text { Precision } \times \text { Recall }}{\beta^{2} \times \text { Precision }+\text { Recall }}
\end{equation}
The parameter $\beta$ is used to control the balance of recall and precision \citep{Lever2016}. When using $\beta = 1$, recall and precision are equally balanced and the formula simplifies to:
\begin{equation}
F_{1} = \frac{2 \times \text { Precision } \times \text { Recall }}{ \text { Precision }+\text { Recall }}
\end{equation}
Since this measure is usually used for binary classification problems and the second subtask is a multi-class problem, a micro-averaging of the scores is needed, which aggregates the individual per-document decisions across all classes to compute the average score \citep{ir-manning}. This averaging method gives equal weight to each classification decision, which leads to a higher impact from the performance of a large class on the results compared to that of a small class. Thus, the micro-averaged results are considered as a measure of the effectiveness on the large classes that can be preferable when dealing with imbalances in class distribution, which is the case when handling the GermEval 2017 datasets. The best results from the GermEval 2017 as well as other publications are reported in Table \ref{results-germeval}.

The winners of the first subtask \citep{Sayyed2017IDS-IUCL:-Inves} on the synchronic test set as well as both subtasks on the diachronic set used word and character n-grams for text representation, in combination with feature selection based on information gain and L1-regularization (Lasso) \citep{Lasso}. In addition, they used adaptive synthetic sampling \citep{4633969} to compensate for imbalances in the class distribution. They performed classification using XGBoost \citep{chen2016}, which is a specific implementation of regularized gradient boosted trees.

Naderalvojoud et al. \cite{Naderalvojoud2017} participated only in the second subtask and achieved the best score on the synchronic test set. They developed a model that utilizes three different German sentiment lexicons that are built using the translation of English lexicons, such as SentiWordNet \citep{esuli-sebastiani-2006-sentiwordnet} and SentiSpin \citep{takamura2005}. In this system, a deep recurrent neural network (RNN) was used to learn contextual sentiment weights and thus to change the polarity of terms depending on the context of their use.

In order to solve the subtasks A and B, Hövelmann \& Friedrich \cite{Hovelmann2017Fasttext-and-Gr} have developed different models and systems. The best model was based on a fastText  classifier \citep{joulin2017bag}, which was enhanced with pre-trained word vectors. This model was able to reach the second best score in both subtasks. Furthermore, a gradient boosted trees (GBT) classifier was developed, which was trained on bag-of-words features combined with the linguistic inquiry and word count (LIWC) \citep{liwc} features. In addition, other models were implemented that used Word2Vec embeddings in combination with classifiers such as GBTs or feedforward multilayer perceptron (MLP). 

Sidarenka \cite{Sidarenka2017PotTS-at-GermEv} developed three systems for the second subtask. The first was an SVM classifier trained on a variety of different features such as character-level features, word-level features, part-of-speech features, and lexicons features. The second system was a bidirectional long short-term memory (Bi-LSTM) \citep{lstm}, which was trained using word embeddings. The last system combined the two systems into an ensemble. This system achieved the third best score out of the 8 participants.

Other participants tried other approaches such as using different lexica or other word embeddings like GloVe instead of Word2Vec or fastText. In addition, some teams used classifiers like conditional random field (CRF) or a stacked learner  \citep{eger-etal-2017-eelection}, which is an ensemble-based method that uses several base classifiers from scikit-learn \citep{sklearn} and a multilayer perceptron as a meta-classifier to combine the predictions of the base classifiers.

After the GermEval 2017 shared task, \cite{10.1007/978-3-030-00794-2_28} conducted experiments using a lexicon-based Bi-LSTM model that yield slightly better results on the sentiment analysis subtask. These results were then outperformed by Biesialska et al. \cite{biesialska-etal-2020-sentiment}, which proposed a transformer-based sentiment analysis (TSA) approach that leverages ELMo contextual embeddings in a model based on the transformer architecture. This approach achieves better results than all reported results for this subtask. In regard to the first subtask, Parcheta et al. \cite{Parcheta2020} experimented using multiple text encoding techniques, such as byte pair encoding (BPE) \citep{10.5555/177910.177914}, GloVe and BERT. To generate the BERT embeddings, they used a small multilingual model that was trained using 104 different languages. The generated embeddings were then used with different architectures, such as a convolutional neural network (CNN) \citep{kim-2014-convolutional}, RNN, LSTM, and gated recurrent units (GRU) \citep{cho-etal-2014-learning}. The variety of the embeddings and architectures resulted in numerous models that performed better than the winning systems of GermEval 2017. The best model uses a combination of BERT and BPE text encoding methods with a single-layered CNN implementation. In a recent work \citep{reeval17}, the authors re-evaluated the GermEval 2017 using pre-trained language models and achieved the best reported scores for both tasks using the German BERT model \textit{bert-base-german-dbmdz-uncased}.

\begin{table}[!ht]
\caption{Best results on the synchronic and diachronic test sets for subtask A on relevance classification and subtask B published in GermEval 2017 and in other works after the competition}
\centering
\begin{adjustbox}{width=1\textwidth}
\begin{tabular}{lcccc}
\hline
\multirow{2}{*}{\textbf{System}} & \multicolumn{2}{c}{\textbf{Subtask A (micro $F1$ ~\%)}} & \multicolumn{2}{c}{\textbf{Subtask B (micro $F1$ ~\%)}} \\ \cline{2-5} 
                                     & Synchronic & Diachronic & Synchronic & Diachronic \\ \hline

XGBoost \citep{Sayyed2017IDS-IUCL:-Inves}                              & \textbf{90.3}      & \textbf{90.6}      & 73.3      & \textbf{75.0}      \\
SWN2-RNN \citep{Naderalvojoud2017}                         & -          & -          & \textbf{74.9}      & 73.6      \\
FastText \citep{Hovelmann2017Fasttext-and-Gr}                            & 89.9      & 89.7      & 74.8      & 74.2      \\ 

Bi-LSTM + SVM \citep{Sidarenka2017PotTS-at-GermEv}   & -      & -      & 74.5      & 71.8      \\ \hline
BERT + PBE + CNN \citep{Parcheta2020}                         & 95.0       & 94.3       & -          & -          \\
ELMo  + TSA \citep{biesialska-etal-2020-sentiment}                          & -          & -          & 78.9           & -     \\
Bert-base-german-dbmdz-uncased \citep{reeval17}                          & \textbf{95.7}          & \textbf{94.8}          & \textbf{80.7}           & \textbf{80.0}     \\\hline
\end{tabular}
\end{adjustbox}
\label{results-germeval}
\end{table}

\section{Proposed Method}
\begin{figure}[!ht]
\centering
\includegraphics[scale=0.45]{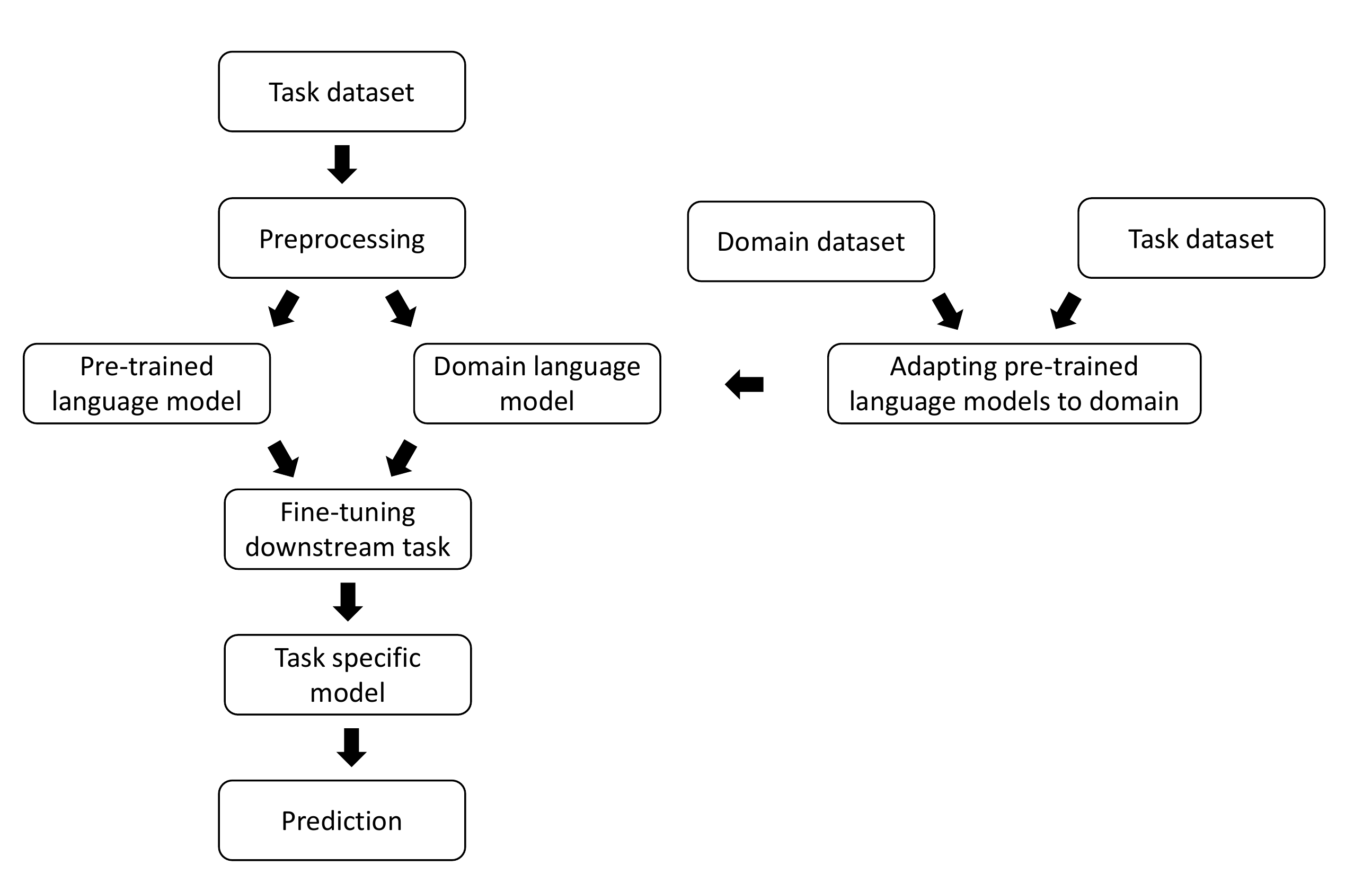}
\caption{Outline of the proposed method}
\label{approach}
\end{figure}

In this section, the suggested method to conduct the experiments will be described, as illustrated in Fig. \ref{approach}. This includes the data preprocessing and the variety of models that are used. 

\label{sec4}

In the first part of this work, transformer-based language models are fine-tuned on the previously mentioned downstream tasks to investigate how they compare to earlier systems. Then, one of the pre-trained models is further trained using masked language modeling on unlabeled domain data, unlabeled task data, and combinations of both to adapt the language models to the specific domain of the downstream tasks. These models are then experimented with and compared to previous results.

\subsection{Preprocessing}
\label{subsec1}
Before starting the process of training models and making predictions, the raw data need to be preprocessed to remove the noise existing in the text. First, duplicates and empty text documents are removed. Then punctuation marks are also deleted. As an exception, repetitions of question marks, exclamation points, and periods are replaced by the terms ``strongquestion'', ``strongexclamation'', and ``annoyeddots''. Furthermore, URLs and numbers got replaced by the terms ``URL'' and ``number''. Whereas, other numerical tokens such as money amounts, dates, and time are replaced by ``money'' and ``dates''. Since many documents originate from Twitter, usernames are replaced by ``twitterusername'' except for the usernames related to the \textit{Deutsche Bahn} like \textit{$@$DB\_Bahn}, \textit{$@$Bahnansagen}, or \textit{$@$Bahn\_Info}, which are pooled by replacing them with ``dbusername''. Additionally, the hashtags mentioned in tweets are modified by removing the ``\#'' character. Words like ``S-Bahn'' and ``S Bahn'' are also combined to the term ``sbahn''. Also, before removing all the punctuation marks, the emoticons ``:('' and ``:-('' are replaced by the token ``sadsmiley'',``:)'', ``:-)'', ``;-)'', ``:-))'' and ``:D'' by ``happysmiley'' and ``:-D'' and ``XD'' by ``laughingsmiley''. For all other possible emoticons, the term ``emote'' is used. Finally, whitespaces and unicode characters like emojis are removed. Excluding the fastText model and the uncased models, no lowercase folding is needed, since all models are trained on cased data. Removing stop words and replacing German umlauts (``ä'', ``Ä'', ``ö'', ``Ö'', ``ü'' and ``Ü'') as well as ligatures (e.g. ``ß'') are briefly tested but did not show any improvements and have not been utilized. Table \ref{examplePre} shows an example of a document before and after applying the mentioned preprocessing techniques. For the domain adaption using masked language modeling, no data preprocessing was applied on the unlabeled data. 

\begin{table}[!ht]
\caption{Example of a document before and after the preprocessing}
\centering
\begin{adjustbox}{width=1\textwidth}
\begin{tabular}{l|l}
\hline
Original & \begin{tabular}[c]{@{}l@{}}@nordschaf theoretisch kannste dir überall im Kölner Stadtbereich was\\ suchen. Mit der KVB + S-Bahn kommt man überall fix hin.\\ (\textbf{Engl.:} @nordschaf theoretically you can look for something anywhere \\in the Cologne city area. With the KVB + S-Bahn you can get\\ everywhere quickly.)\end{tabular} \\ \hline
Preprocessed & \begin{tabular}[c]{@{}l@{}}twitterusername theoretisch kannste dir überall im Kölner Stadtbereich \\was suchen Mit der KVB sbahn kommt man überall fix hin \end{tabular}\\ \hline
\end{tabular}%
\end{adjustbox}
\label{examplePre}
\end{table}

\subsection{System Description}
\label{subsec2}
In the attempt to reach high scores in the first two subtasks, different systems and models were used. For the additional pre-training and fine-tuning of transformer-based models, an NVIDIA A100 GPU was used. The pre-training time usually varied between two and three hours, whereas the fine-tuning took around 8 minutes for the base models and 20 minutes for the large ones. 

As a baseline, a fastText classifier \citep{joulin2017bag} was trained on the preprocessed text. This classifier is constructed with the goal to predict a class or a label instead of a word, which is the case when using an unsupervised algorithm like continuous bag-of-words (CBOW) to generate word embeddings. In addition to word embeddings, fastText also uses character-level n-grams, which makes it capable of handling morphologically rich languages like German, and sentences with a variety of words. Deviating from the default configuration, the dimensionality of the word vectors was set to 50, the learning rate was initialized with 0.1. For the loss computation, softmax was used, and the number of word n-grams was set to 4. The classifier was then trained for 20 epochs. To ensure the reproducibility of the results, the number of used threads was set to 1. As an additional model, these parameters and the collected tweets were used to generate word embeddings, which then were utilized with the fastText classifier. 

To further improve results, transformer-based models were used, starting with BERT \citep{bert_devlin}. This model was designed to pre-train deep bidirectional representations from an unlabeled text by jointly exploring left and right contexts on all levels. As a result, the pre-trained BERT language model can be fine-tuned with only one additional output layer to develop models that achieve remarkable results for different tasks. The authors published two models: BERT$_{Base}$ and BERT$_{Large}$. Details of these models are shown in Table \ref{details}. In addition, they also released a multilingual model, which was trained on cased data in 104 languages.

These models were pre-trained in two phases: ``masked language modeling'', and ``next sentence prediction''. In the first phase, instead of predicting every next token, the model only predicts a percentage of random ``masked'' words from a sentence. The second phase is a binary classification task in which the model predicts whether the second sentence is the actual next sentence of the first sentence. Due to the computational expense of training such models from scratch, all used models are already pre-trained by other organizations and were made publicly available. The first German model was published by the German company Deepset AI\footnote{\url{https://deepset.ai/german-bert} (last access: 10-04-2022)}. It was trained on German Wikipedia dump, court decisions, and news articles. The \textit{Digitale Bibliothek Münchener
Digitalisierungszentrum} (DBMDZ\footnote{\url{https://github.com/dbmdz/berts} (last access 10-04-2022)}) released two additional German models, cased and uncased. These were trained on German Wikipedia dump, European Union Bookshop corpus, OpenSubtitles, and Web Crawls. The two teams have joined forces and released two new BERT models (GBERT$_{Base}$ and GBERT$_{Base}$) \citep{chan-etal-2020-germans}, which outperform the previously released models and were trained on four different datasets: the German portion of the \textit{Open Super-large Crawled ALMAnaCH coRpus} (OSCAR) \citep{ortiz-suarez-etal-2020-monolingual}, German Wikipedia dump, \textit{The Open Parallel Corpus} (OPUS) \citep{tiedemann-2012-parallel}, and Open Legal Data \citep{10.1145/3383583.3398616}. Additional information about these models and the other models used to conduct the experiments are listed in Table \ref{models-corpus}.

After the release of BERT, a Robustly Optimized BERT pre-training approach (RoBERTa) \citep{liu2020roberta} was introduced. It enhances the BERT approach by changing the pre-training procedure through training the model longer, over more data, on longer sequences, and by removing the next sentence prediction and using improved hyperparameters. For this work, GottBERT \citep{scheible2020gottbert} was utilized, which is a German RoBERTa$_{Base}$ model that was trained on the German part of OSCAR data.

\begin{table}[!ht]
\caption{Details of the different types of transformer-based models}
\centering
\begin{adjustbox}{width=1\textwidth}
\begin{tabular}{l|c|c|c|c|c|c}
\hline
 & $\textbf{BERT}_{\textbf{Base}}$ & $\textbf{BERT}_{\textbf{Large}}$ & $\textbf{RoBERTa}_{\textbf{Base}}$ & $\textbf{XLM RoBERTa}_{\textbf{Large}}$ & $\textbf{ELECTRA}_{\textbf{Base}}$ & $\textbf{ELECTRA}_{\textbf{Large}}$
\\[4pt] \hline
Layers & 12 & 24 & 12 & 24 & 12 & 24 \\ 
Attention heads & 12 & 16 & 12 & 16 & 12 & 16 \\ 
Hidden states & 768 & 1024 & 768 & 1024 & 768 & 1024 \\ 
Parameters & 110M & 336M & 125M & 550M & 110M & 335M \\ \hline
\end{tabular}
\end{adjustbox}
\label{details}
\end{table}

In addition to these models, the transformer-based model XLM RoBERTa \citep{DBLP:conf/acl/ConneauKGCWGGOZ20} is used, which is a cross-lingual language model that was trained on 2.5TB of data across 100 languages. XLM RoBERTa outperforms previous multilingual approaches by incorporating more training data and languages, including low-resource languages. Although multiple XLM RoBERTa models are available, only the large model was considered for this work.

Another tested model is ELECTRA (Efficiently Learning an Encoder that Classifies Token Replacements Accurately) \citep{clark2020electra}. This model introduced a different approach to language pre-training, where it uses another task called replaced token detection (RTD). Instead of masking the input by replacing some words with the token ``[MASK]'' as in BERT, ELECTRA corrupts the input tokens by replacing them with synthetically generated tokens. It then trains the model to distinguish between ``real'' and ``fake'' input data.  This is achieved using a discriminator that classifies the tokens and a generator which provides plausible fake tokens. Both  transformer-based components are trained jointly. In addition to the two BERT models, Chan et al. \cite{chan-etal-2020-germans} also released two German ELECTRA models: a base model and a large one. In their benchmarking, the large model reached state-of-the-art performance in three downstream tasks.

To adapt the language models to the task domain, multiple experiments were conducted on GBERT$_{Large}$. These experiments were based on continuing pre-training using masked language modeling on the collected tweets and on combinations of the unlabeled task data and parts of the unlabeled tweets. Additional experiments were conducted by expanding the vocabulary of the pre-trained model with around 20k new words from the unlabeled tweets, where the embeddings of the new tokens are initialized randomly. Moreover, for some experiments instead of masking 15~\% of the text during the masked language modeling, 30~\% was masked, since based on a recent study \citep{Wettig2022ShouldYM} masking up more than 15~\% of the tokens can be beneficial in some cases.

\begin{table}[!ht]
\caption{Details about the used pre-trained models}
\centering
\begin{adjustbox}{width=1\textwidth}
\begin{tabular}{l|l|c|c|c}
\hline
\textbf{Model} & \textbf{Type} & \textbf{\#Languages} & \textbf{Data size} & \textbf{Data source}                                                               \\ \hline
GBERT          & BERT          & 1                    & 163.4GB            & \begin{tabular}[c]{@{}c@{}}OSCAR, OPUS, Wikipedia, \\ Open Legal Data\end{tabular} \\
mBERT          & BERT          & 104                  & unknown            & Wikipedia                                                                          \\
DBMDZ BERT & BERT & 1 & 16GB & \begin{tabular}[c]{@{}c@{}}Wikipedia, EU Bookshop corpus, \\ Open Subtitles, Common-, Para-, \\ NewsCrawl\end{tabular} \\
XLM RoBERTa    & RoBERTa       & 100                  & 2.5T               & CommonCrawl, Wikipedia                                                             \\
GottBERT       & RoBERTa       & 1                    & 145GB              & OSCAR                                                                              \\
GELECTRA       & ELECTRA       & 1                    & 163.4GB            & \begin{tabular}[c]{@{}c@{}}OSCAR, OPUS, Wikipedia, \\ Open Legal Data\end{tabular} \\ \hline
\end{tabular}%
\end{adjustbox}
\label{models-corpus}
\end{table}

For the baseline classifier, the fastText\footnote{\url{https://github.com/facebookresearch/fastText} (last access: 19-04-2022)} (version 0.9.1) was used. The training and fine-tuning of transformer-based models were conducted using the Transformers \citep{wolf-etal-2020-transformers} library by HuggingFace\footnote{\url{https://github.com/huggingface/transformers} (last access: 19-04-2022)}. For all models, the same hyperparameters were used (see Table \ref{hype}). In addition, the continued pre-training of the models on the unlabeled domain and task data was performed for 5 epochs using the same hyperparameters except the max sequence length, which was set to 512. 

\begin{table}[!ht]
\caption{Hyperparameters used for the transformer-based models}
\centering
\begin{tabular}{l|l}
\hline
\textbf{Hyperparameter} & \textbf{Value} \\ \hline
Optimizer & AdamW \citep{AdamW} \\
Learning rate & 2e-5 \\
Manual seed & 42 \\
Max sequence length & 256 \\
Warmup steps & 100 \\
Adam epsilon & 2e-8 \\
Epochs & 5 \\ 
Batch size & 32\\\hline
\end{tabular}
\label{hype}
\end{table}

\section{Results and Discussion}
\label{sec5}

Table \ref{cross} shows the results of 5-fold cross-validation for both subtasks on the training set. All average scores were obtained using 5-fold cross-validation and are complemented by their respective standard deviation. 

\begin{table}[!ht]
\caption{Results of the average score and standard deviation using 5-fold cross-validation on the training set for subtask A on relevance classification and subtask B on sentiment detection. (*) denotes p value < 0.05 in a Wilcoxon test against the fastText model. The best results are highlighted in bold}
\centering
\begin{adjustbox}{width=1\textwidth}
\begin{tabular}{lcc}
\hline
\textbf{System}                      & \textbf{Subtask A (micro $F1$ ~\%)} & \textbf{Subtask B (micro $F1$ ~\%)} \\ \hline
DBMDZ BERT$_{Base}$ cased                & 94.6 $\pm$ 0.3*               & 81.0 $\pm$ 0.3*               \\
DBMDZ BERT$_{Base}$ uncased              & 94.9 $\pm$ 0.1*               & 81.5 $\pm$ 0.3*               \\
GBERT$_{Base}$                           & 94.6 $\pm$ 0.1*               & 81.7 $\pm$ 0.3*                \\
GBERT$_{Large}$                          & 95.2 $\pm$ 0.2*               & 84.4 $\pm$ 0.4*               \\
GottBERT$_{Base}$                        & 94.6 $\pm$ 0.1*                & 82.1 $\pm$ 0.4*               \\
mBERT cased                          & 93.4 $\pm$ 0.3*               & 79.3 $\pm$ 0.4*               \\
XLM RoBERTa$_{Large}$                    & 94.8 $\pm$ 0.3*               & 83.1 $\pm$ 0.4*               \\
GELECTRA$_{Base}$                        & 94.4 $\pm$ 0.2*               & 81.0 $\pm$ 0.5*               \\
GELECTRA$_{Large}$                       & 95.0 $\pm$ 0.2*               & 84.2 $\pm$ 0.4*               \\ \hline
GBERT$_{Large}$ + Domain         & 95.2 $\pm$ 0.2*               & 84.3 $\pm$ 0.4*               \\
GBERT$_{Large}$ + Domain + 30~\% Mask        & 95.2 $\pm$ 0.4*               & \textbf{84.8 $\pm$ 0.4}*               \\
GBERT$_{Large}$ + Domain + Vocab           & 95.1 $\pm$ 0.3*               & 83.8 $\pm$ 0.2*               \\
GBERT$_{Large}$ + Domain + Vocab + 30~\% Mask             & 94.9 $\pm$ 0.2*               & 83.8 $\pm$ 0.3*               \\
GBERT$_{Large}$ + Task    & 95.1 $\pm$ 0.1*               & 84.2 $\pm$ 0.4*               \\
GBERT$_{Large}$ + Task + 30~\% Mask           & 95.3 $\pm$ 0.2*               & 84.2 $\pm$ 0.3*               \\
GBERT$_{Large}$ + Task + Domain (100K)          & \textbf{95.3 $\pm$ 0.2}*               &84.3 $\pm$ 0.3*               \\
GBERT$_{Large}$ + Task + Domain (100K) + 30~\% Mask            & 95.3 $\pm$ 0.1*               & 84.4 $\pm$ 0.3*          \\
GBERT$_{Large}$ + Task  + Domain (200K)                & 95.3 $\pm$ 0.1*                & 84.6 $\pm$ 0.2*                \\
GBERT$_{Large}$ + Task + Domain (200K) + 30~\% Mask  & 95.1 $\pm$ 0.1*              & 84.8 $\pm$ 0.6*               \\ \hline
FastText (baseline)                  & 90.5 $\pm$ 0.5               & 77.6 $\pm$ 0.7\\
FastText + Domain (1M) embeddings    & 90.8 $\pm$ 0.4               & 78.4 $\pm$ 0.5               \\ \hline
\end{tabular}
\end{adjustbox}
\label{cross}
\caption{}
\end{table}

To analyze the results of the cross-validation and assess how the systems perform against the fastText model, which can be considered a strong baseline, a one-sided Wilcoxon rank-sum test \citep{10.2307/3001968} was used. The tests were conducted using the statistical programming language R (version 4.1.2) \citep{rstudio} with the significance level $\alpha = 0.05$. 

Since all $p$-values of the conducted tests are below the significance level, the null hypothesis can be rejected, which indicates that all systems outperform the fastText model. 

Table \ref{results} shows the obtained results on the test datasets for both subtasks using systems trained on the training and development sets. Based on the scores, all transformer-based systems outperform the baseline model. For the first subtask, off-the-shelf GBERT$_{Large}$ improved the results obtained by the winning system from GermEval 2017 by about +5.6 percentage points micro-averaged $F1$-Score on the synchronic test set, and by +4.7 percentage points on the diachronic test set. It also outperforms the best reported scores \citep{Parcheta2020} by +0.2 percentage points and +0.5 percentage points, respectively. The obtained scores of this model were even slightly improved after a second pre-training phase using additional domain data, with the best scores reaching 96.1~\% and 95.9~\%. Out of the off-the-shelf models, GBERT$_{Large}$ and GELECTRA$_{Large}$ were also able to reach the best scores on the second subtask. GBERT$_{Large}$ improves upon the best GermEval systems by a +8.8 percentage points margin on the first test set, whereas GELECTRA$_{Large}$ reached +9.3 percentage points improvement on the second one. Compared to the best score in \citep{reeval17}, these models reached a score that is +3.0 percentage points better on the synchronic set and +4.3 on the diachronic set. Similar to the results of the first subtask, the highest scores (85.1~\% and 85.3~\%) were reached when a continued pre-training on the domain and task data was applied.     

These results indicate that continuing to pre-train language models on domain-specific unlabeled data as well as the task data usually improves results, which was the case for GBERT$_{Large}$ as shown in Table \ref{results}. Based on the results, increasing the masking percentage can also improve the results, but it is not always the case. Furthermore, adding new domain words to the vocabulary of the pre-trained model did not show any improvements and sometimes led to slightly worse results than the off-the-shelf pre-trained models, which can be a result of the random initialization of the embeddings. Although this method showed some improvements, it can probably be more beneficial when using higher quality data that has been properly selected and filtered. 

Also noteworthy, despite XLM RoBERTa being a multilingual model, it outperforms most of the German Base models on both subtasks.

\begin{table}[!ht]
\caption{Results on the synchronic and diachronic test sets of the different systems trained on the training and development datasets for subtask A on relevance classification and subtask B on sentiment detection, as well as comparison with results from other publications. The best results are highlighted in bold.}
\centering
\begin{adjustbox}{width=1\textwidth}
\begin{tabular}{lcccc}
\hline
\multirow{2}{*}{\textbf{System}} & \multicolumn{2}{c}{\textbf{Subtask A (micro $F1$ ~\%)}} & \multicolumn{2}{c}{\textbf{Subtask B (micro $F1$ ~\%)}} \\ \cline{2-5} 
                                     & Synchronic & Diachronic & Synchronic & Diachronic \\ \hline
DBMDZ BERT$_{Base}$ cased                & 95.6      & 95.4      & 81.1      & 78.3      \\
DBMDZ BERT$_{Base}$ uncased              & 95.4      & 95.3      & 81.0      & 80.4      \\
GBERT$_{Base}$                           & 94.5      & 94.9      & 82.2      & 80.7      \\
GBERT$_{Large}$                          & 95.9      & 95.3      & 83.7      & 82.9       \\
GottBERT$_{Base}$                        & 94.5       & 94.9       & 82.2      & 81.7      \\
mBERT cased                          & 94.0      & 93.7      & 78.4      & 76.7       \\
XLM RoBERTa$_{Large}$                    & 95.6      & 95.0      & 82.1       & 81.8      \\
GELECTRA$_{base}$                        & 94.7      & 93.8      & 81.5       & 79.4      \\
GELECTRA$_{Large}$                       & 95.6      & 95.6       & 83.3      & 84.3      \\ \hline
GBERT$_{Large}$ + Domain         & 95.9      & 95.4      & 84.2      & 84.8      \\
GBERT$_{Large}$ + Domain + 30~\% Mask         & 95.5      & 95.7      & \textbf{85.1}      & 84.0      \\
GBERT$_{Large}$ + Domain + Vocab          & 95.7      & 94.8      & 83.1      & 82.9      \\
GBERT$_{Large}$ + Domain + Vocab + 30~\% Mask                & 95.5      & 94.8      & 82.9      & 83.3      \\
GBERT$_{Large}$ + Task & 95.8      & \textbf{95.9}      & 84.0      & \textbf{85.3}      \\
GBERT$_{Large}$ + Task + 30~\% Mask            & 95.9      & 95.4       & 84.6       & 84.4      \\
GBERT$_{Large}$ + Task + Domain (100K)          & 95.8      & 95.6      & 83.8      & 84.4      \\
GBERT$_{Large}$ + Task + Domain (100K) + 30~\% Mask             & 96.1      & 95.9      & 84.0      & 85.1      \\
GBERT$_{Large}$ + Task + Domain (200K               & 95.6      & 95.8      & 84.1      & 85.0      \\
GBERT$_{Large}$ + Task + Domain (200K) + 30~\% Mask  & \textbf{96.1}      & 95.1      & 84.5       & 84.4      \\ \hline
bert-base-german-dbmdz-uncased \citep{reeval17}      & 95.7      & 94.8      & 80.7      & 80.0  \\
BERT + PBE + CNN \citep{Parcheta2020}                         & 95.0       & 94.3       & -          & -          \\
ELMo  + TSA \citep{biesialska-etal-2020-sentiment}                          & -          & -          & 78.9           & -     \\
XGBoost \citep{Sayyed2017IDS-IUCL:-Inves}                              & 90.3      & 90.6      & 73.3      & 75.0      \\
SWN2-RNN \citep{Naderalvojoud2017}                         & -          & -          & 74.9      & 73.6      \\
fastText \citep{Hovelmann2017Fasttext-and-Gr}                            & 89.9      & 89.7      & 74.8      & 74.2      \\
 \hline
fastText (baseline)                  & 90.7      & 89.6      & 75.8      & 74.9      \\
fastText + Domain (1M) embeddings    & 90.6      & 90.2      & 75.9      & 75.2      \\ \hline
\end{tabular}
\end{adjustbox}
\label{results}
\end{table}

Although the transformer-based models showed that they can reach remarkable results, one of their disadvantages is that using them is computationally expensive and can require specific hardware such as GPUs, which can become costly when used in production at scale.  Moreover, training these models take considerably more time in comparison to models such as fastText, which usually just needs a couple of minutes only using one CPU. Besides these disadvantages, there are also some challenges and limitations when dealing with sentiment analysis in social networks. Determining sentiment in tweets for example, where the text is usually coupled with hashtags, emojis, and links can be very difficult \citep{7543720}. Additionally, analyzing a textual expression from a semantic point of view can be crucial to detecting the underlying sentiment \citep{POZZI20171}. This is usually not taken into consideration when dealing with sentiment analysis, where a sentence is taken just as it is, which can result in wrong interpretations. Furthermore, using a word or phrase that entails an intentional deviation from its literal definition can hinder detecting the correct expressed sentiment. This is the case when dealing with sarcasm and irony, which are usually difficult to recognize - not only for machines, but also for humans. This difficulty can result in poor performance even for state-of-the-art systems. In addition, collecting data can also be challenging since searching for a specific term can result in collecting unwanted data. For example, the term \textit{bahn} does not only refer to the train in German but can also refer to a track or anything that can be laid in straight lines, such as \textit{Laufbahn} (\textbf{Engl.} running track) \citep{germevaltask2017}. This problem can be avoided if the data is carefully and properly filtered beforehand.
Another limitation, which needs to be considered, is the possibility that the provided annotations are not entirely correct and that a number of the predictions are assumed wrong even though they are actually correct. An example for this case is shown in Table \ref{tab:example-annot}, where the document can be considered positive but was mistakenly annotated as negative and therefore the prediction is treated as a wrong classification.

\begin{table}[!ht]
\caption{Example for a poorly annotated document}
\centering
\resizebox{\textwidth}{!}{%
\begin{tabular}{l|l|l}
\hline
Document &
  Annotation &
  Prediction \\ \hline
\begin{tabular}[c]{@{}l@{}}Heute bin ich mal wieder eine Runde \#U5 gefahren. Ging trotz Streik bei \\ der S-Bahn gut voran \#weilwirdichlieben http://t.co/aa2lFkDQNT\\ (\textbf{Engl.:} Today I rode again a round \#U5. Went well despite strike at the \\ S-Bahn \#becauseweloveyou http://t.co/aa2lFkDQNT)\end{tabular} &
  Negative &
  Positive \\ \hline
\end{tabular}%
}
\label{tab:example-annot}
\end{table}

\section{Conclusion}
\label{sec6}

Transformer-based architectures such as BERT, ELECTRA, RoBERTa, and XLM RoBERTa were used in this work for the first two subtasks of GermEval 2017. They showed a remarkable improvement over the results that were reached when the competition was held, as well as those that were reported by subsequent works. Moreover, the conducted experiments revealed that adapting these pre-trained models to the domain using unlabeled task and domain data can even outperform the achieved results. 

The findings show that continuing pre-training language models using task data and domain-specific unlabeled data is an interesting concept to consider whenever the initial language model data is non-specific for the intended use. Such improvements are the reason why these models have gained massive popularity in recent years in different NLP downstream tasks. This success is due to the improvement of context understanding that heavily benefits from pre-training huge language models on enormous amounts of data. Although the results showed using domain-specific data lead to improvements upon off-the-shelf models, an extended analysis of the used data is needed. Such an analysis can help determine how to best use domain-specific data for additional pre-training and how the quality of the data can improve the models.

In addition to the already mentioned models, other transformer-based models such as ALBERT, XLNet, DeBERTa, and T5 were also released. Unfortunately, for most of these, no German pre-trained models are available. In future works, these models can also be tested to investigate how they compare with the reported systems, assuming that in the future more German models are going to be released. On the other hand, there is already evidence that multilingual and novel transformer architectures perform similarly well as language-specific and domain-specific language models under certain conditions. Thus, a comparison of novel multilingual architectures, such as modular transformers \citep{pfeiffer-etal-2022-lifting}, with the results of domain-adapted models is also useful for future work.

\section*{Declarations}
\subsection*{Funding}
The work of Ahmad Idrissi-Yaghir and Henning Schäfer was funded by a PhD grant from the DFG Research Training Group 2535 Knowledge- and data-based personalisation of medicine at the point of care (WisPerMed).

\subsection*{Conflict of interest}
The authors declare that they have no competing interests.

\subsection*{Availability of Data and Materials}
Data used in preparation of this article is publicly available and was obtained from the GermEval 2017 Shared Task\footnote{\url{https://sites.google.com/view/germeval2017-absa/} (last access: 17-05-2022)}. 

\subsection*{Authors’ Contributions}
The conceptualization of the study was carried out by CMF and AIY. CMF and AIY planned the experiments. AIY and HS implemented the software, executed the experiments and analysed the data. AIY has written the original draft under the supervision of CMF and NB. All authors read and approved the final manuscript.





\bibliography{sn-bibliography}


\end{document}